\documentclass[journal abbreviation, manuscript]{copernicus}
\usepackage{graphicx}%
\usepackage{multirow}%
\usepackage{amsmath,amssymb,amsfonts}%
\usepackage{amsthm}%
\usepackage{mathrsfs}%
\usepackage[title]{appendix}%
\usepackage{xcolor}%
\usepackage{textcomp}%
\usepackage{manyfoot}%
\usepackage{booktabs}%
\usepackage{algorithm} 
\usepackage{algorithmic}  
\usepackage[algo2e]{algorithm2e} 
\usepackage{listings}%
\usepackage{comment}%
\usepackage[most]{tcolorbox}
\usepackage{subcaption}

\begin{document}
\nolinenumbers

\title{ChatEarthNet: A Global-Scale Image-Text Dataset Empowering Vision-Language Geo-Foundation Models}


\Author[][]{Zhenghang}{Yuan} 
\Author[]{Zhitong}{Xiong}
\Author[]{Lichao}{Mou}
\Author[]{Xiao Xiang}{Zhu}
\affil[]{Data Science in Earth Observation, Technical University of Munich, Munich 80333, Germany}




\runningtitle{TEXT}

\runningauthor{TEXT}

\received{}
\pubdiscuss{} 
\revised{}
\accepted{}
\published{}


\firstpage{1}

\maketitle

\begin{abstract}
An in-depth comprehension of global land cover is essential in Earth observation, forming the foundation for a multitude of applications. Although remote sensing technology has advanced rapidly, leading to a proliferation of satellite imagery, the inherent complexity of these images often makes them difficult for non-expert users to understand. Natural language, as a carrier of human knowledge, can be a bridge between common users and complicated satellite imagery. In this context, we introduce a global-scale, high-quality image-text dataset for remote sensing, providing natural language descriptions for Sentinel-2 data to facilitate the understanding of satellite imagery for common users. Specifically, we utilize Sentinel-2 data for its global coverage as the foundational image source, employing semantic segmentation labels from the European Space Agency's (ESA) WorldCover project to enrich the descriptions of land covers. By conducting in-depth semantic analysis, we formulate detailed prompts to elicit rich descriptions from ChatGPT. To enhance the dataset's quality, we introduce the manual verification process. This step involves manual inspection and correction to refine the dataset, thus significantly improving its accuracy and quality. Finally, we offer the community ChatEarthNet, a large-scale image-text dataset characterized by global coverage, high quality, wide-ranging diversity, and detailed descriptions. ChatEarthNet consists of 163,488 image-text pairs with captions generated by ChatGPT-3.5 and an additional 10,000 image-text pairs with captions generated by ChatGPT-4V(ision). This dataset has significant potential for training vision-language geo-foundation models and evaluating large vision-language models for remote sensing. The dataset will be made publicly available.

\end{abstract}

\begin{figure}
\centering
\includegraphics[width=0.85\textwidth]{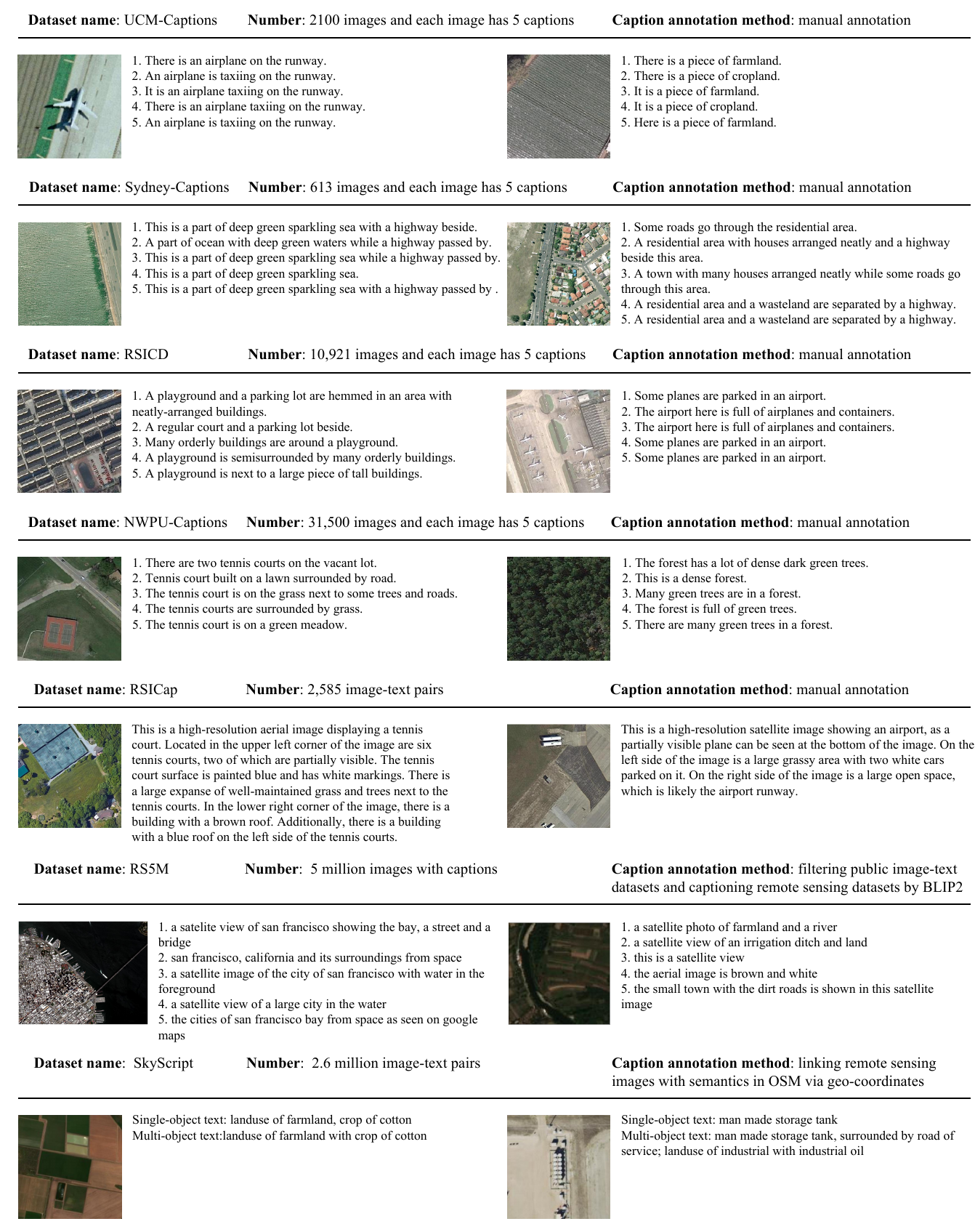}
\caption{Comparative Visualization of Image-Text Pairs across UCM-Captions \citep{qu2016deep}, Sydney-Captions \citep{qu2016deep}, RSICD \citep{lu2017exploring}, NWPU-Captions \citep{cheng2022nwpu}, RSICap \citep{hu2023rsgpt}, RS5M \citep{zhang2023rs5m}, SkyScript \citep{wang2023skyscript} Datasets.}\label{existing_data}
\end{figure}

\introduction  
Land cover refers to the surface components of land such as the water body, tree, bare land, developed area, etc., providing the landscape patterns and features on the Earth's surface. A comprehensive understanding of global land cover is of importance for international initiatives, such as the United Nations Framework Convention on Climate Change (UNFCCC) \citep{mora2014global}, as well as various applications, including urban planning, environmental assessment, disaster response, and economic development \citep{garcia2012land}. Satellite imagery in the field of remote sensing is regarded as the ideal data for land cover monitoring, as it can provide an overview and repetitive observations of land cover \citep{franklin2002remote}. The Sentinel-2 mission \citep{drusch2012sentinel} has achieved great success in providing comprehensive satellite images that enable the monitoring of the Earth's surface on a global scale. A thorough analysis of land cover using Sentinel-2 data not only enhances the understanding of terrestrial ecosystems but also supports numerous practical applications including natural resource management, agriculture and food security \citep{Sentinel-2}.

Despite the rapid advancements in remote sensing technology, which leads to an exponential increase in satellite imagery \citep{xiong2022earthnets}, the inherent complexity embedded in these satellite images often makes them difficult for common users to understand. Natural language, with its rich semantic information, is regarded as a bridge between common users and complicated satellite imagery, serving as a crucial modality for understanding sophisticated machine learning systems \citep{lobry2021rsvqa}. For example, natural language is integrated into machine learning systems of different tasks in a user-friendly manner, such as image captioning \citep{lu2017exploring}, visual question answering \citep{lobry2020rsvqa, yuan2022easy}, visual grounding \citep{wen2023vision, zhan2023rsvg}, and referring image segmentation \citep{yuan2023rrsis} in the remote sensing domain. 
In this context, Vision-Language Models (VLMs) \citep{kuckreja2023geochat} are proposed to learn vision-language correlation from image-text pairs, where text is used to describe the image. This represents a significant step forward for VLMs, showcasing their versatile capability through zero-shot transfer across various tasks \citep{zhang2023vision, liu2023remoteclip}. However, this versatile capability is largely attributed to their use of extensive, semantically aligned image-text pair datasets.

Datasets play an important role in driving advancements of deep learning-based VLMs \citep{silva2024large, kuckreja2023geochat}, which require large amounts of image-text pair data to learn the parameters of models. VLMs can utilize web-scale image-text pairs available on the internet, where images are associated with corresponding relevant text. However, few pairs on the website provide detailed descriptions for satellite images \citep{wang2023skyscript}. This further confirms the demand to construct large-scale and high-quality image-text datasets for remote sensing.

Figure \ref{existing_data} shows the comparative visualization along with the number and caption (description) annotation methods of existing available image-text pair datasets in remote sensing domain, including UCM-Captions \citep{qu2016deep}, Sydney-Captions \citep{qu2016deep}, RSICD \citep{lu2017exploring}, NWPU-Captions \citep{cheng2022nwpu}, RSICap \citep{hu2023rsgpt}, RS5M \citep{zhang2023rs5m}, SkyScript \citep{wang2023skyscript} Datasets. These datasets range significantly in size, quality of caption, and annotation method. The variation in dataset sizes from thousands to millions of image-text pairs. While the RS5M and SkyScript datasets are generated automatically through algorithms, allowing them to 
reach the million level in terms of quantity, their text descriptions lack detail and only provide very basic information. Similarly, smaller datasets like UCM-Captions, Sydney-Captions, RSICD, and NWPU-Captions predominantly feature simple captions, often limited to a single sentence for each caption. Though five captions are provided per image, the descriptions tend to be very similar or even identical. This simplicity and redundancy are disadvantages of these datasets. RSICap dataset stands out for its detailed manual annotation, but the quantity is limited, with only 2,585 image-text pairs. This is because it is manually annotated, a time and labour consuming process, making large-scale dataset generation difficult. In conclusion, these datasets suffer from limitations, with none of them encompassing both a large quantity with global coverage and high-quality descriptions. It is necessary to create an image-text dataset with high quantity and quality.

Our motivation is to construct a large-scale image-text dataset with global coverage that not only meets the semantic richness required for advanced VLMs training but also extends the understanding of satellite imagery to common users. For the data sources, we utilize Sentinel-2 data due to its practicality and accessibility. For the source of semantic information in Sentinel-2 data, we choose land cover maps from the European Space Agency's (ESA) WorldCover project \citep{WorldCover}. Leveraging Sentinel-2 data and its land cover map, we aim to construct a global-scale, high-quality image-text dataset, which is essential for training robust VLMs. However, it is challenging to manually annotate Sentinel-2 data on a large scale with high quality. This is mainly because manually annotating large datasets is time and labour consuming, and the low resolution of Sentinel-2 images makes it challenging to distinguish land covers. 

In this study, we introduce an automated processing framework for generating descriptions of satellite images, leveraging the powerful language generation capability of ChatGPT \citep{ChatGPT}. This framework, through the design of effective prompts, can make use of ChatGPT to yield high-quality and detailed descriptions for Sentinel-2 imagery at a global scale. By integrating rich natural language descriptions with global satellite imagery, the proposed dataset fills in the interpretability gap between complex satellite imagery and common users. To further improve the quality of the dataset, we conduct a manual validation process to check the caption's correctness and quality. In summary,  we offer the community ChatEarthNet, a large-scale image-text dataset with global coverage, high quality, wide-ranging diversity, and detailed descriptions. This dataset has great potential for training foundation models.

Foundation models, pre-trained on extensive datasets, can be further fine-tuned for specific tasks across different domains, serving as versatile tools in artificial intelligence \citep{zhou2023comprehensive}. However, vision-language geo-foundation models face challenges due to the scarcity of high-quality, geotagged linguistic and visual data. In this scenario, ChatEarthNet emerges as a vital resource. It provides high-quality, geotagged language and image data required for training vision-language geo-foundational models and evaluating the performance of large vision-language models in remote sensing applications.


\begin{figure}[h]%
\centering
\includegraphics[width=0.97\textwidth]{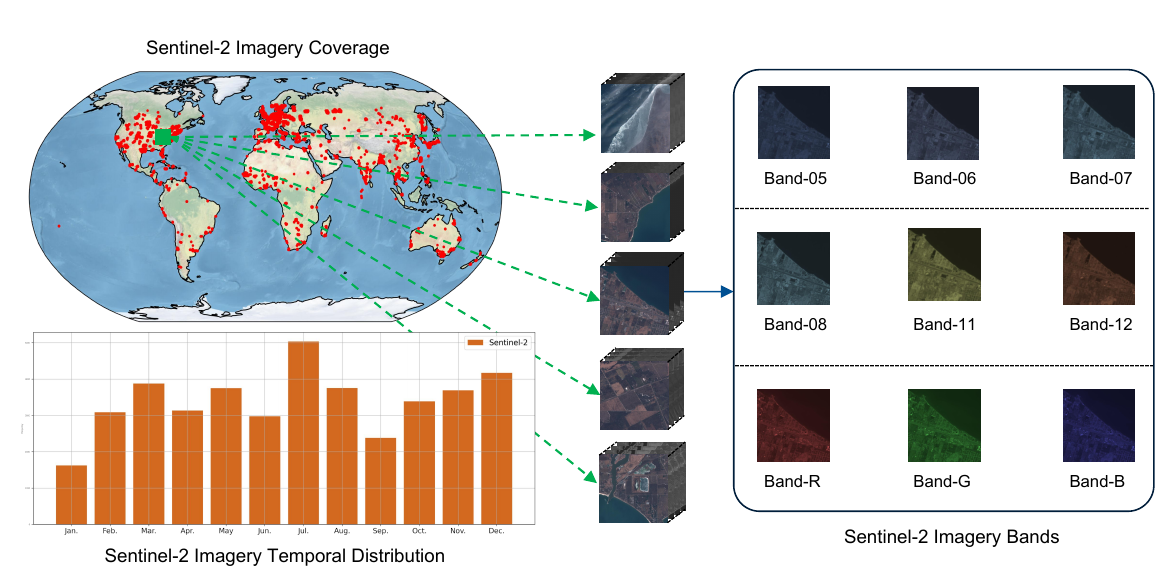}
\caption{The upper-left part of the figure displays the geographical distribution of the Sentinel-2 data used in the ChatEarthNet dataset. The lower-left part shows the temporal distribution of the Sentinel-2 data used. The right part visualizes some examples of the images and the nine spectral bands used in the dataset.}\label{S2_vis}
\end{figure}

\section{Dataset and Methodology}
ChatEarthNet dataset is built upon the Sentinel-2 data \citep{drusch2012sentinel} with global coverage and the fine-grained land cover product from the ESA's WorldCover project \citep{WorldCover}. The overall dataset construction process is that for each Sentinel-2 image, we analyze its land cover type distributions and design sophisticated prompts to generate descriptive texts using ChatGPT. This approach ensures that each description accurately reflects the visual data, providing a rich semantic description of the satellite imagery. Manual verification and correction of generated texts further improve the dataset's accuracy and quality.

\subsection{Sentinel-2 Data in ChatEarthNet}

We follow the sampling strategy utilized in the SatlasPretrain dataset \citep{bastani2023satlaspretrain}. In the dataset construction, we source images from the Sentinel-2 collected in SatlasPretrain for further developing the image-text dataset. This subsection details the characteristics of Sentinel-2 data \citep{Sentinel_Online} used in the ChatEarthNet dataset.

\begin{enumerate} 
\item \textbf{Global Distribution}: ChatEarthNet dataset is designed to capture a detailed description of the land cover, with its images spanning all continents except Antarctica and encompassing major urban centers, shown in the upper-left part of Figure \ref{S2_vis}. The global distribution ensures diverse landscapes and urban areas, enriching the dataset with a variety of visual characteristics relevant to different geographical locations.

\item \textbf{Temporal Coverage}: The temporal distribution of images is a critical aspect of the dataset. As illustrated in the bottom-left side of Figure \ref{S2_vis}, the ChatEarthNet dataset includes Sentinel-2 images captured throughout different months to ensure that they cover different seasons of the Earth's surface. This temporal diversity allows the dataset to provide a more comprehensive appearance of different land cover types.

\item \textbf{Image Size}: The spatial size of Sentinel-2 images in the ChatEarthNet dataset is 256x256 pixels. There are a total of 163,488 images in the dataset, providing extensive coverage across the world and enabling analysis and applications in various remote sensing tasks.

\item \textbf{Spectral Band}: Sentinel-2 imagery is rich in spectral information, and the ChatEarthNet dataset includes nine specific bands from the S2A sensor, as shown in the right part of Figure \ref{S2_vis}. These bands are band 5, band 6, band 7, band 8, band 11, band 12, along with the red, green, and blue (RGB) bands. The selected bands offer a detailed spectral resolution that captures a broad range of wavelengths, providing insights into different physical properties of the land cover. 

\begin{figure}[h]%
\centering
\includegraphics[width=0.99\textwidth]{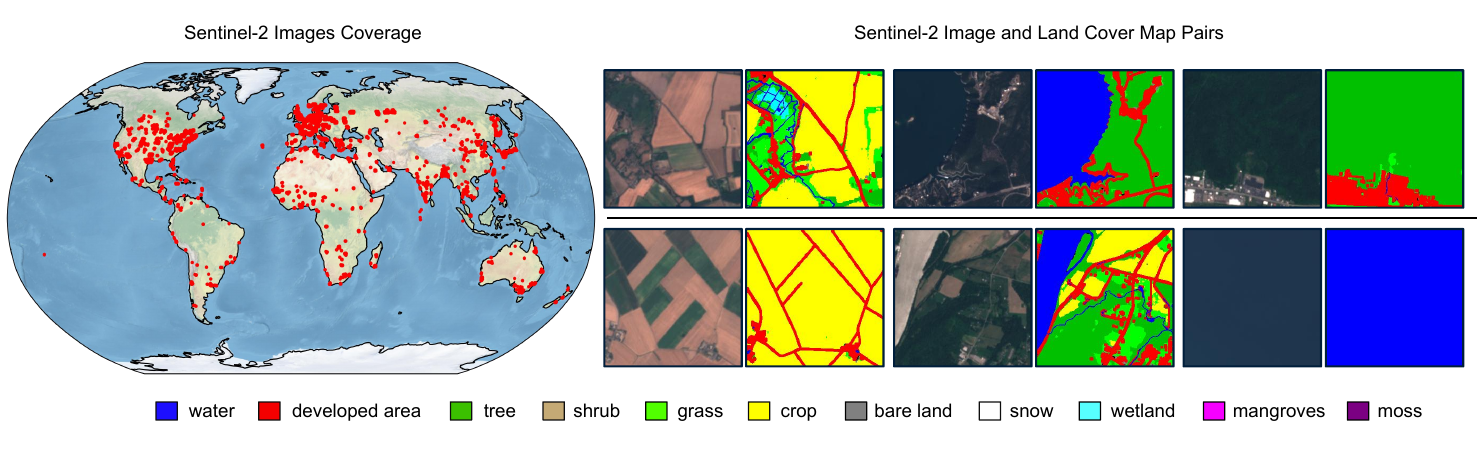}
\caption{The left side of the figure depicts the geographical distribution of land cover maps sourced from the WorldCover product and utilized in the ChatEarthNet dataset. On the right side, examples of segmentation maps showcase various land cover types.}\label{LC_vis}
\end{figure}

\end{enumerate}

\subsection{Land Cover Map from WorldCover Product}

As depicted in Figure \ref{LC_vis}, the ChatEarthNet dataset leverages the WorldCover product 2020 version \citep{WorldCover} to obtain semantic information, which provides global-scale land cover maps. These maps describe the land cover at 10 meters resolution. Specifically, we utilize 11 different land cover classes \citep{bastani2023satlaspretrain}, including ``water'', ``developed area'', ``tree'', ``shrub'', ``grass'', ``crop'', ``bare land'', ``snow'', ``wetland'', ``mangroves'', and ``moss''. These land cover types offer a detailed categorization that encompasses natural and urban-related landscapes, providing critical semantic information to generate text. The integration of the WorldCover product with Sentinel-2 data provides a robust foundation for our image-text dataset. By doing so, not only global-scale satellite images but also detailed land cover semantics are provided.

\subsection{Prompt Design}

In this section, we detail the prompt designs for the caption generation.
Numerous Large Language Models (LLMs) have been developed, among which ChatGPT is distinguished by its exceptional performance. Its proficiency in textual understanding and production makes it a valuable tool for textual analysis and description generation. Thus, in this work, we adopt two LLMs, ChatGPT-3.5 and ChatGPT-4V(ision) to generate two different versions of image-text datasets.

While both ChatGPT-3.5 and ChatGPT-4V represent significant advancements in the field of LLMs, they exhibit differences in their performance and capabilities. Compared with ChatGPT-3.5, ChatGPT-4V is a multimodal LLM, which can process not only text but also visual inputs, thereby enhancing its contextual comprehension of the shapes and spatial distributions of land cover types in images. Moreover, ChatGPT-4V also demonstrates improved performance in terms of accuracy, coherence, and the ability to handle nuanced prompts. However, considering the Application Programming Interface (API) prices, ChatGPT-4V is much more expensive than ChatGPT-3.5. Additionally, as of now, for usage tier 1, ChatGPT-4V has a limit of 500 requests per day, while ChatGPT-3.5 has a limit of 10,000 requests per day. This means that if processing one image (image represented by text for GPT-3.5) requires a single request, GPT-3.5 can handle 10,000 images per day, while GPT-4V is limited to processing just 500 images daily. Considering cost and efficiency, we utilize ChatGPT-3.5 for generating descriptions for the complete dataset, comprising 163,488 image-text pairs, and randomly select a subset of 10,000 Sentinel-2 images for description generation using ChatGPT-4V, resulting in 10,000 image-text pairs.

Given that ChatGPT is predominantly trained on commonly available natural images, its direct application to satellite images may not yield optimal results. To address this, we carefully design prompts that embed semantic information from the land cover maps for the caption generation. This allows the language models to utilize the provided context to generate precise and semantically rich descriptions for satellite images. Although the underlying concept is straightforward, it requires meticulous designs to compensate inherent limitations of current ChatGPT. These limitations include challenges with instructions following, where LLMs may not strictly follow the given instructions in the prompt. The other limitation is the well-known hallucination problem, where LLMs may output plausible but factually incorrect or nonsensical information. To alleviate these issues, we meticulously design the prompts and instructions to guide ChatGPT toward generating reliable and contextually appropriate descriptions. 

The term ``prompt'' in LLMs like ChatGPT refers to the input provided to the model to generate a response. System prompt and user prompt serve different functions, as illustrated below:
\begin{enumerate}
    \item \textbf{System Prompt}: System prompt typically refers to the initial instructions set by developers for configuring LLMs. Its purpose is to establish the ground rules or guidelines for the following conversation, including setting the tone, style, or scope of the responses to standardize the LLM's behavior.
    \item \textbf{User Prompt}: User prompt is the actual question or statement input by the user, seeking a response from LLMs. It's the variable part of the interaction that can differ with each user. Context information can be part of the user prompt to provide background details necessary for LLMs to generate relevant and coherent responses. It's additional information to give LLMs a better understanding of the current topic.
\end{enumerate}

\subsubsection{Prompt Design for ChatGPT-3.5}
We aim to generate captions, i.e., natural language descriptions, for Sentinel-2 satellite images. To provide sufficient semantic information, we leverage the geographic-aligned land cover maps derived from the WorldCover product. Given the corresponding land cover map, we generate textual descriptions based on the proportions of different land covers. Since ChatGPT-3.5 can only accept text instructions as input, we need to extract the semantic information from land cover maps and provide it to ChatGPT-3.5 in the textual form. The designed prompt for ChatGPT-3.5 is presented in the following block. 
 
\begin{tcolorbox}[colback=blue!5!white,colframe=blue!75!black,title=Prompt Design for ChatGPT-3.5]
\label{prompt35}
{messages} = [{"Analyze the provided image as an AI visual assistant. The following contexts are provided. The overall land cover distributions from most to least are: <\textcolor{red}{Algorithm 1}>
You are an AI visual assistant who can help describe images based on the given contexts. Please write the description in a paragraph, and avoid saying other things. The following constraints should be obeyed:
\begin{enumerate}
    \item Describe the image in the order of the spatial distributions presented in the given contexts. Link descriptions of different parts to make the overall image description more fluent.
    \item Describe the dominant land cover type in the image and its spatial locations.
    \item Describe the land cover types in each part of the image in descending order of their coverage areas.
    \item Diversify descriptions related to portions in each paragraph. 
    \item Summarize the main theme of the image in the final sentence.
    \item Describe it objectively; do not use words: `possibly', `likely', `perhaps', `context', `segmentation', `appear', `change', `transition', `dynamic', or any words with similar connotations."]
\end{enumerate}
}
\end{tcolorbox}

\begin{algorithm}[H]
\DontPrintSemicolon
\KwData{Land cover map $Y$}
\KwResult{Generated prompt containing land cover proportions for $Y$}
\BlankLine
\SetKwFunction{FMain}{GeneratePrompt}
\SetKwProg{Fn}{Function}{:}{}
\Fn{\FMain{$Y$}}{
    \, Compute the overall proportions of different land cover types in $Y$ \\
    Generate a prompt describing the overall land cover proportions \\
    Split $Y$ into five patches: $Y_{tl}$, $Y_{tr}$, $Y_{bl}$, $Y_{br}$, $Y_{m}$ \\
   \ForEach{patch $Y_i$}{
        \, Compute the number of land cover types $n_i$ in $Y_i$ \\
       Compute the proportion of each land cover type in the patch\\
       Sort the land cover types according to proportions in a descending order\\
       Select three (if $n_i$ is less than three, select $n_i$) land cover types from most to least \\
       Generate a prompt for the patch describing the portions of all land cover types
    }
   Concatenate all prompts \\
   \KwRet concatenated prompts\
}
\caption{Generating the prompt for land cover proportion}
\label{A1}
\end{algorithm} 

The prompt comprises two elements: the system prompt to guide the response style and set constraints to ChatGPT, and the user prompt containing context derived from land cover maps using \textbf{Alrogithm}\ref{A1}. The system prompt includes a set of explicit constraints to ensure the generated descriptions are fluent, accurate, and unbiased. Specifically, we force ChatGPT to generate fluent descriptions and focus more on the spatial locations and portions of different land cover types. We also encourage ChatGPT to describe objectively and avoid the use of subjective words. For the user prompt, we extract the semantic information from land cover maps, where each pixel represents a land cover type. Specifically, in \textbf{Alrogithm}\ref{A1}, initially, we calculate the overall proportions of different land cover types and generate a prompt describing the overall land cover proportions. Subsequently, we split the land cover map $Y \in \mathbb{R}^{256\times 256}$ into four non-overlapping patches of equal size, each being $128\times 128$. The patches are defined from the origin of the map, which starts at the top-left corner (0,0) and spans to (255,255). The top-left patch, denoted as $Y_{tl} \in \mathbb{R}^{128\times 128}$, extends from indices 0 to 127 in both row and column directions. The top-right ($Y_{tr}$), bottom-left ($Y_{bl}$), and bottom-right ($Y_{br}$) quadrants are similarly demarcated. Additionally, we extract a middle patch $Y_{m}$, also $128\times 128$ in size, centered within the map, with indices ranging from 64 to 191 in both the row and column directions, aligning with the midpoint of the land cover map. For each patch $Y_i$, we calculate the proportion of each land cover within that patch relative to its total pixel count. We then rank these land cover types by their proportional presence and select the top three to represent the primary land cover types of the patch. In cases where a patch contains fewer than three land cover types, we select all available types. This selection process is employed because ChatGPT-3.5 tends to generate verbose descriptions when presented with abundant data. Limiting the information to three main land cover types ensures more focused descriptions, avoiding unnecessarily lengthy descriptions. After determining the primary land cover types for all five patches in a land cover map $Y$, we concatenate their proportions and the overall proportions to formulate the final prompt. This tailored prompt enables ChatGPT-3.5 to generate accurate, detailed, and coherent descriptions of Sentinel-2 satellite imagery.

\subsubsection{Prompt Design for ChatGPT-4V}
The prompt design for ChatGPT-4V is presented in the following block. Similarly, this prompt also contains the system prompt and the user prompt. However, the system prompt for ChatGPT-4V differs from that used for ChatGPT-3.5, as ChatGPT-4V is capable of processing the land cover map as an image directly. Given that the land cover map is essentially a segmentation map where each color represents a land cover type, this key information is provided to ChatGPT-4V through the system prompt. To enhance the accuracy and detail of descriptions, we also define several guides and constraints in the system prompt. Moreover, considering the API request limit of ChatGPT-4V, we put four images into one request to generate descriptions more efficiently. While ChatGPT-4V can handle image inputs, it still requires specific guidance to accurately interpret segmentation maps from a remote sensing perspective. Hence, the user prompt is supplemented with semantic information extracted from the land cover maps using \textbf{Algorithm} \ref{A2} and \textbf{Algorithm} \ref{A3}.

\begin{tcolorbox}[colback=blue!5!white,colframe=blue!75!black,title=Prompt Design for ChatGPT-4V]
messages = [{ "You are an AI visual assistant that can analyze the given image. In the image, {different colors represent different land cover types}. The color for the land cover dictionary is: `[0, 0, 255] (blue): {water}; [255, 0, 0](red): {developed area}; [0, 192, 0] (dark green): {tree}; [200, 170, 120] (brown): {shrub}; [0, 255, 0] (green): {grass}; [255, 255, 0] (yellow): {crop}; [128, 128, 128] (grey): {bare}; [255, 255, 255] (white): {snow}; [0, 255, 255] (cyan): {wetland}; [255, 0, 255] (pink): {mangroves}; [128, 0, 128] (purple): {moss}.' You will be provided with four independent images at once.

For the first/second/third/fourth images, the distribution of each land cover type is:
<\textcolor{red}{Algorithm 2}>
For the first/second/third/fourth images, the spatial distribution of the image is:
<\textcolor{red}{Algorithm 3}>
You are given four independent images, describe in long sentences for each image separately using four paragraphs, and avoid saying other things. The following constraints should be obeyed: 
\begin{enumerate}
    \item Do not use color-related words; treat the color as the land cover type directly.
    \item Generate the four descriptions separately; do not add connections between them. 
    \item When describing water, developed, and crop areas, incorporate shape descriptors.
    \item Double-check all the presented land cover types based on the distribution of each land cover type. If some land covers are not presented, do not mention them.
    \item Describe it objectively; do not use words: `possibly', `likely', `perhaps', `color dictionary', `appear', `change', `transition', `dynamic', or any words with similar connotations.
    \item Double-check the shape and location of the developed area, water course, grass, tree, shrub, wetland, and crop areas based on the given image if they are present.
    \item Consider the spatial statistics as a unified image without breaking them down into individual spatial distributions and land cover proportions when describing the overall scene.
    \item Describe each land cover separately for each given image, and then describe the main theme of each given image."]
\end{enumerate}}
\end{tcolorbox}

Similar to the process described in \textbf{Alrogithm}\ref{A1}, we split the land cover map $Y \in \mathbb{R}^{256\times 256}$ into five different patches: top-left $Y_{tl}$, top-right $Y_{tr}$, bottom-left $Y_{bl}$, bottom-right $Y_{br}$ and middle $Y_{m}$ patches, each being $128\times 128$. As shown in \textbf{Algorithm} \ref{A2}, for each patch, we calculate the proportion of each land cover within that patch relative to its total pixel count. Different from \textbf{Alrogithm}\ref{A1}, we provide the proportion information of all land cover types (instead of three main land cover types) in each patch to ChatGPT-4V. The reason is that ChatGPT-4V is more powerful and can process all information to generate detailed descriptions without unnecessarily lengthy descriptions. In \textbf{Algorithm} \ref{A3}, we aim to calculate the distribution of each land cover type across the five patches. For each land cover type $L_j$, we first calculate the number of pixels for $L_j$ in the land cover map $Y$, represented by $N_j$. Subsequently, for each patch $Y_i$, we calculate the pixel count of $L_j$ in $Y_i$, denoted as $n_{ji}$. The spatial distribution is evaluated using the ratio $\frac{n_{ji}}{N_j}$, which quantifies the presence of $L_j$ in each patch relative to its overall occurrence. After computing the spatial distribution of $L_j$ across all patches, we concatenate prompts for all land cover types. These prompts, derived from calculations in both algorithms, are put into the final text prompt. This text prompt and the land cover map as visual input are then provided to ChatGPT-4V to generate descriptions.

\begin{algorithm}[H]
\DontPrintSemicolon
\KwData{Land cover map $Y$}
\KwResult{Generated prompt containing land cover proportions for five patches of $Y$}
\BlankLine
\SetKwFunction{FMain}{GeneratePrompt}
\SetKwProg{Fn}{Function}{:}{}
\Fn{\FMain{$Y$}}{
    Split $Y$ into five patches: $Y_{tl}$, $Y_{tr}$, $Y_{bl}$, $Y_{br}$, $Y_{m}$ \\
    \ForEach{patch $Y_i$}{
        \, Compute the proportion of each land cover type in the patch \\
        Generate a prompt for the patch describing the portions of all land cover types
    }
    \, Concatenate prompts for all patches\\
    \KwRet concatenated prompts\
}
\caption{Generating the prompt for land cover proportion in each patch}
\label{A2}
\end{algorithm} 

\begin{algorithm}[H]
\caption{Generating the prompt for the spatial distribution of each land cover type}
\label{A3}
\KwData{Land cover map: $Y$}
\KwResult{Spatial distribution of each land cover type across patches.}
\BlankLine
\SetKwFunction{FMain}{GeneratePrompt}
\SetKwProg{Fn}{Function}{:}{}
\Fn{\FMain{$Y$}}{
Split $Y$ into five patches: $Y_{tl}$, $Y_{tr}$, $Y_{bl}$, $Y_{br}$, $Y_{m}$ \\
\ForEach{land cover type $L_j$}{
    \, Calculate the number of pixels for land cover type $L_j$ in $Y$, denoted as $N_j$ \\  
    \ForEach{patch $Y_i$}{
        \, Calculate the number of pixels for land cover type $L_j$ in $Y_i$, denoted as $n_{ji}$ \\
        Calculate the spatial distribution via $\frac{n_{ji}}{N_j}$ \
    }     
    \, Generate a prompt describing the spatial distribution of $L_j$ across all patches. \\
Concatenate prompts for all land cover types
}
\KwRet concatenated prompts\
}
\end{algorithm}

\subsection{Manual verification}
To further improve the quality of the dataset, we conduct a manual validation process to check the caption's correctness and quality. For efficiency and cost savings, ChatGPT-4V involves inputting four images per request in caption generation. Despite the prompt requesting to ``Generate the four descriptions separately; do not add connections between them'', some captions still contain comparisons between the four images. We therefore manually check all captions and refine comparison-related captions. ChatGPT-3.5 processes a single image (represented by text) per request, avoiding comparison issues. We manually inspected 10,000 image-text pairs from ChatGPT-3.5-generated captions, finding no significant issues.

\section{Dataset Analysis and Discussion}
In this section, we present a comprehensive analysis of the ChatEarthNet dataset from different aspects. As we construct the dataset using ChatGPT-3.5 and ChatGPT-4V, we analyze and compare these two different versions to provide a clear overview and understanding.

\begin{figure}[h]%
\centering
\includegraphics[width=0.99\textwidth]{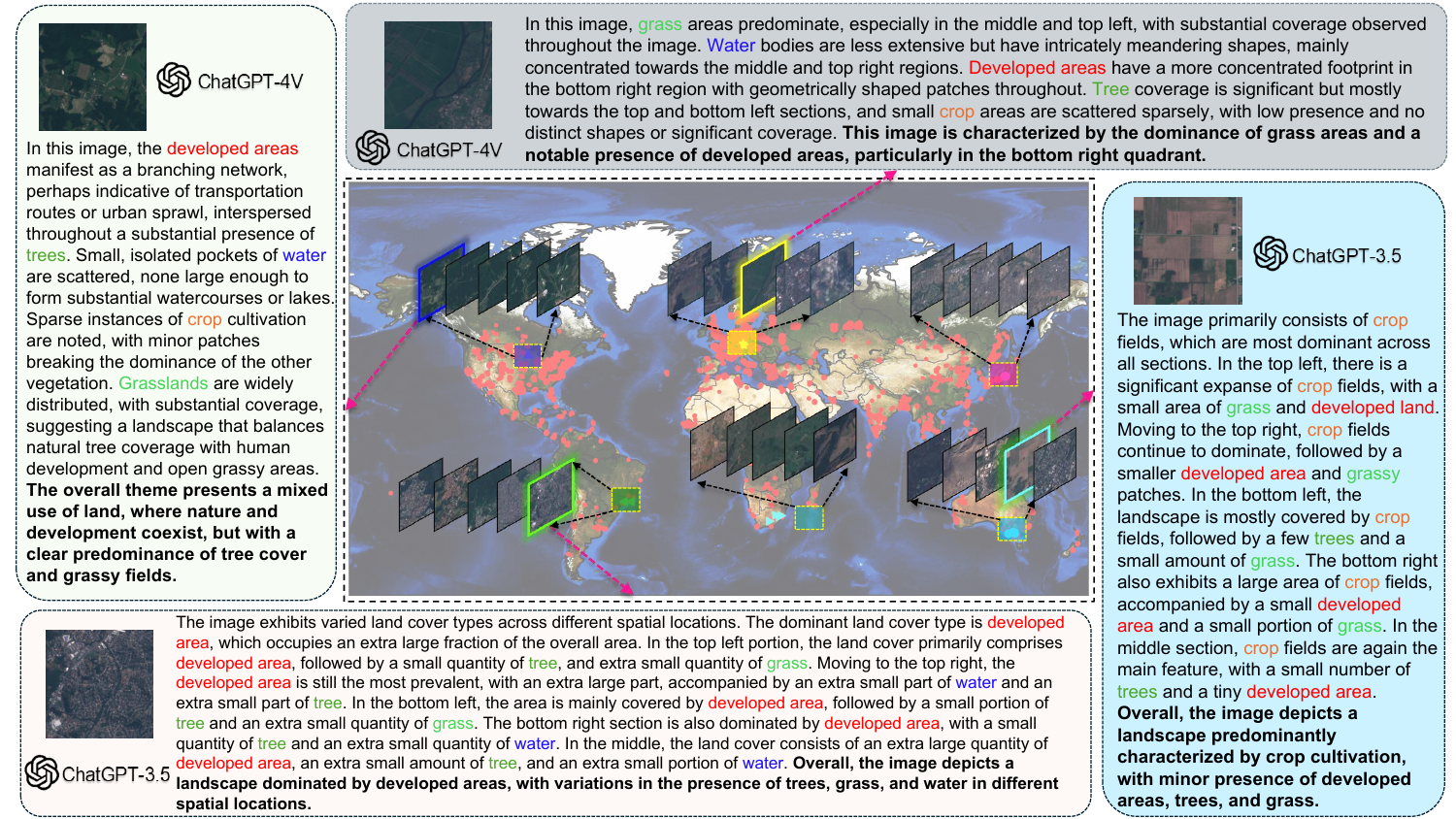}
\caption{An overview of the ChatEarthNet dataset. We randomly select image-text samples from four different locations. The left and top sides display the descriptions generated by ChatGPT-4V. While the right and bottom sides show two samples produced by ChatGPT-3.5. We use different colors to highlight the words of different land cover types.}\label{overall_fig}
\end{figure}
\subsection{Dataset overview}

In Figure \ref{overall_fig}, we present four different image-text pairs from four regions of the Earth, illustrating that images from different geographical locations exhibit unique characteristics. The diversity in land cover distributions across these images is evident. The accompanying texts accurately reflect the quantity and spatial distribution of the various land cover types observed.

In Table \ref{overall}, we present the number of Sentinel-2 images used for generating captions, along with the corresponding numbers of captions generated by ChatGPT-3.5 and ChatGPT-4V in the ChatEarthNet dataset. Specifically, we use 163,488 Sentinel-2 images and generate a long caption accompanying each image using ChatGPT-3.5. For the ChatGPT-4V version, we randomly select 10,000 Sentinel-2 images across the world and generate one detailed caption for each image. In terms of the number of image-text pairs, the ChatEarthNet dataset is not the largest, but it stands out as the first dataset offering high-quality detailed land cover descriptions on a global scale. This makes it a solid foundation for training multi-modal vision-language models in the field of remote sensing.

\begin{table}[h!]
\centering
\caption{The number of Sentinel-2 images used for generating captions, along with the corresponding numbers of captions generated by ChatGPT-3.5 and ChatGPT-4V.}
\begin{tabular}{lccc}
\hline
Subsets & Number of Sentinel-2 Images & Number of ChatGPT-3.5 Captions & Number of ChatGPT4-V Captions \\
        & (ChatGPT-3.5/ChatGPT-4V)    &                                  &                                  \\
\hline
Train   & 98,092/6000                  & 98,092                           & 6000                             \\
Val     & 16,348/1000                  & 16,348                           & 1000                             \\
Test    & 49,048/3000                  & 49,048                           & 3000                             \\
Sum     & 163,488/10,000               & 163,488                          & 10,000                           \\
\hline
\end{tabular}
\label{overall}
\end{table}

\subsection{Geographic coverage}
\begin{figure}
  \begin{minipage}[t]{0.5\linewidth}
    \centering
    \includegraphics[scale=0.17]{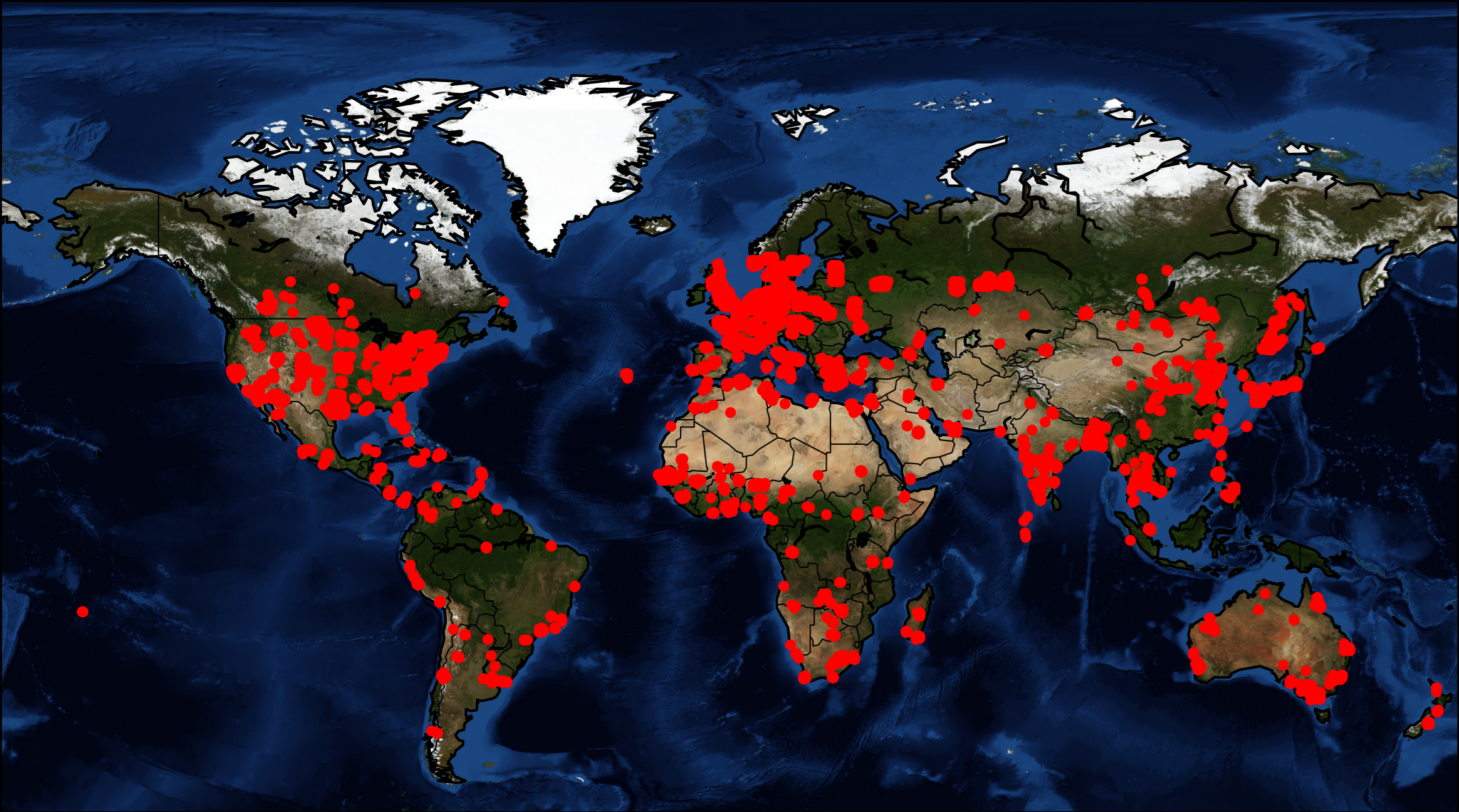}
    \caption{Geographical distribution of image-text pairs using \\ ChatGPT-3.5}
    \label{gpt3.5_2}
  \end{minipage}%
  \begin{minipage}[t]{0.5\linewidth}
    \centering
    \includegraphics[scale=0.17]{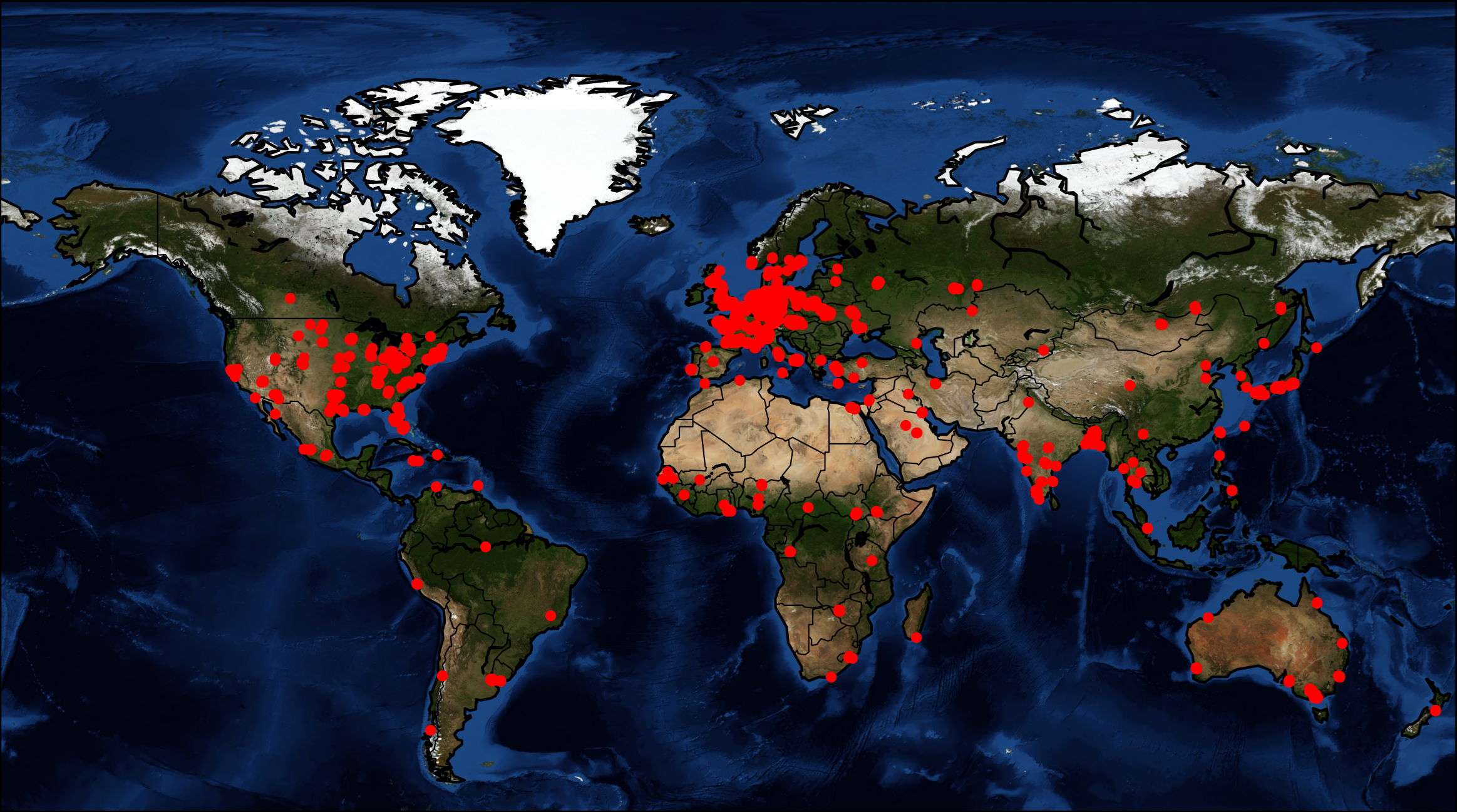}
    \caption{Geographical distribution of image-text pairs using \\ ChatGPT-4V.}
    \label{gpt4_2}
  \end{minipage}
\end{figure}

Figure \ref{gpt3.5_2} and Figure \ref{gpt4_2} illustrate the geographical distribution of image-text pairs using ChatGPT-3.5 and ChatGPT-4V in the ChatEarthNet dataset, respectively. From the two figures, we can see that the image-text pairs for both the ChatGPT-3.5 and ChatGPT-4V versions cover all continents except Antarctica. Compared to the image-text pairs using ChatGPT-4V, the geographical distribution of those using ChatGPT-3.5 is more dense, covering a wider range of areas. Nevertheless, 10,000 high-quality image-text pairs using  ChatGPT-4V are sufficient for fine-tuning LLMs.

\subsection{Word Frequency}

\begin{figure}
  \begin{minipage}[t]{0.5\linewidth}
    \centering
    \includegraphics[scale=0.2]{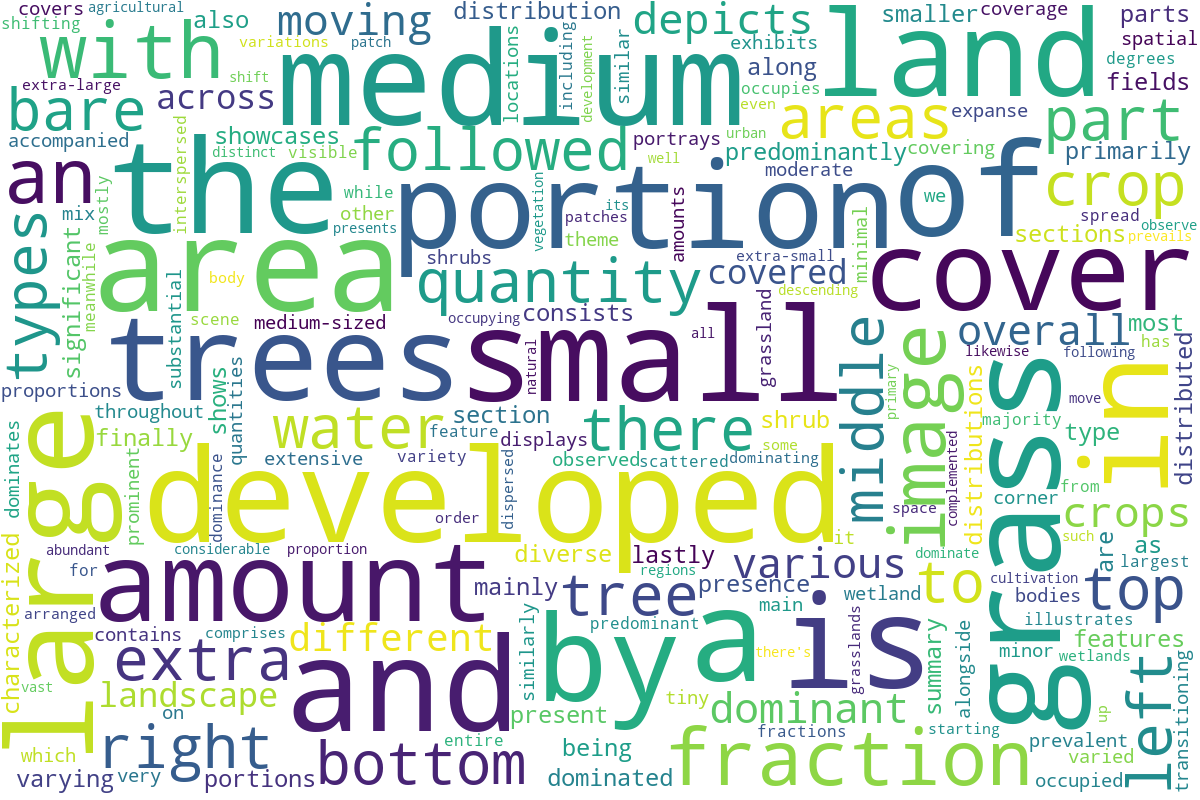}
    \caption{Word cloud for captions generated by ChatGPT-3.5.}
    \label{WC1}
  \end{minipage}%
  \begin{minipage}[t]{0.5\linewidth}
    \centering
    \includegraphics[scale=0.2]{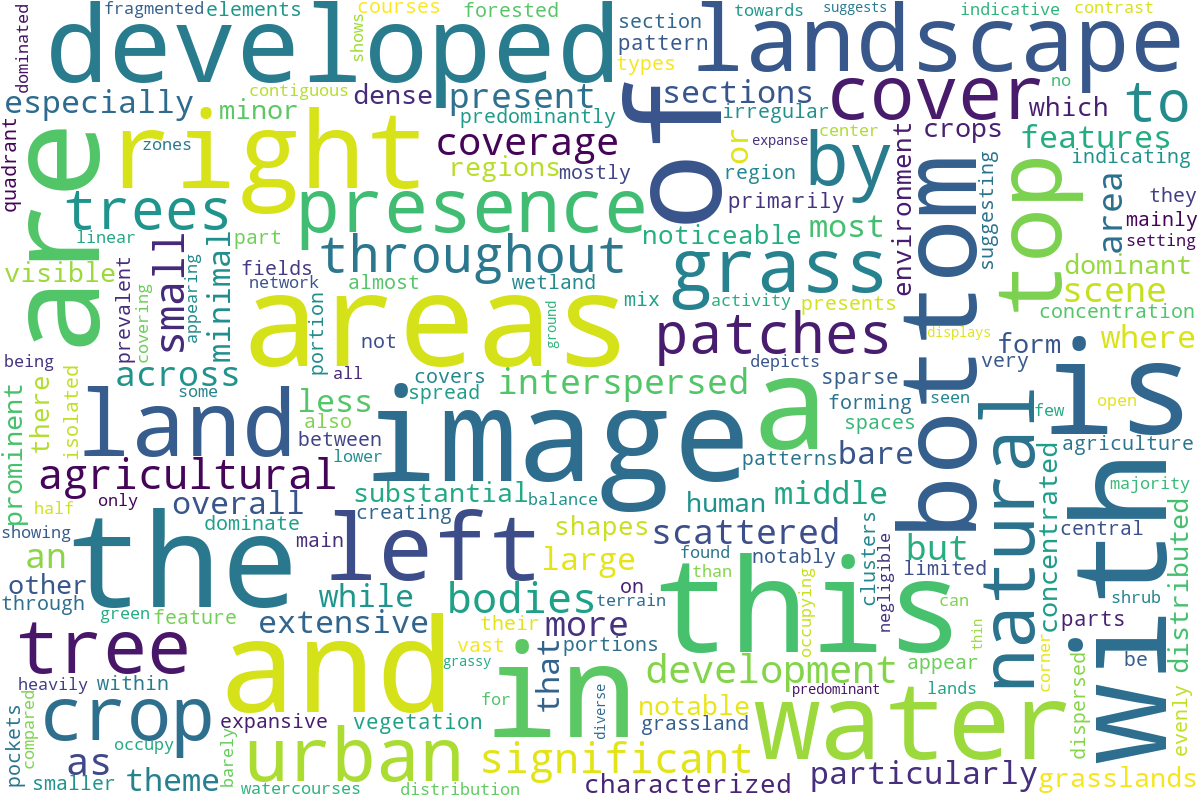}
    \caption{Word cloud for captions generated by ChatGPT-4V.}
    \label{WC2}
  \end{minipage}
\end{figure}

Figure \ref{WC1} and Figure \ref{WC2} illustrate the word clouds for captions generated by ChatGPT-3.5 and ChatGPT-4V, respectively. In the two figures, larger words indicate a higher frequency of occurrence. In Figure \ref{WC1}, prominent words like ``developed'',  ``small'',  ``medium'', ``grass'', and ``portion'' indicates a focus on describing the content and scale of land covers. Other significant words like ``right'' and ``bottom'' relate to specific locations in the image. In Figure \ref{WC2}, the word cloud centers around ``image'' and ``areas'', indicating these are key themes in the generated captions. Adjacent to these are other significant words like ``developed'', ``bottom'', ``water'', ``right'' and ``landscape'', suggesting an emphasis on geographical features and the layout in the image. Words such as ``urban'', ``crop'' and ``natural'' are also prominent, denoting the content involving descriptions of human activities, agricultural, and natural landscapes. In addition, the words ``significant'', ``small'', and ``large'' describe different sizes of land covers. Overall, the captions generated by ChatGPT-3.5 provide more straightforward descriptions focusing on the distribution and size of land cover types. The captions generated by ChatGPT-4V use more varied and descriptive language and showcase a more diverse vocabulary to describe scale of land covers and their layout.

\begin{figure}[h]%
\centering
\includegraphics[width=0.99\textwidth]{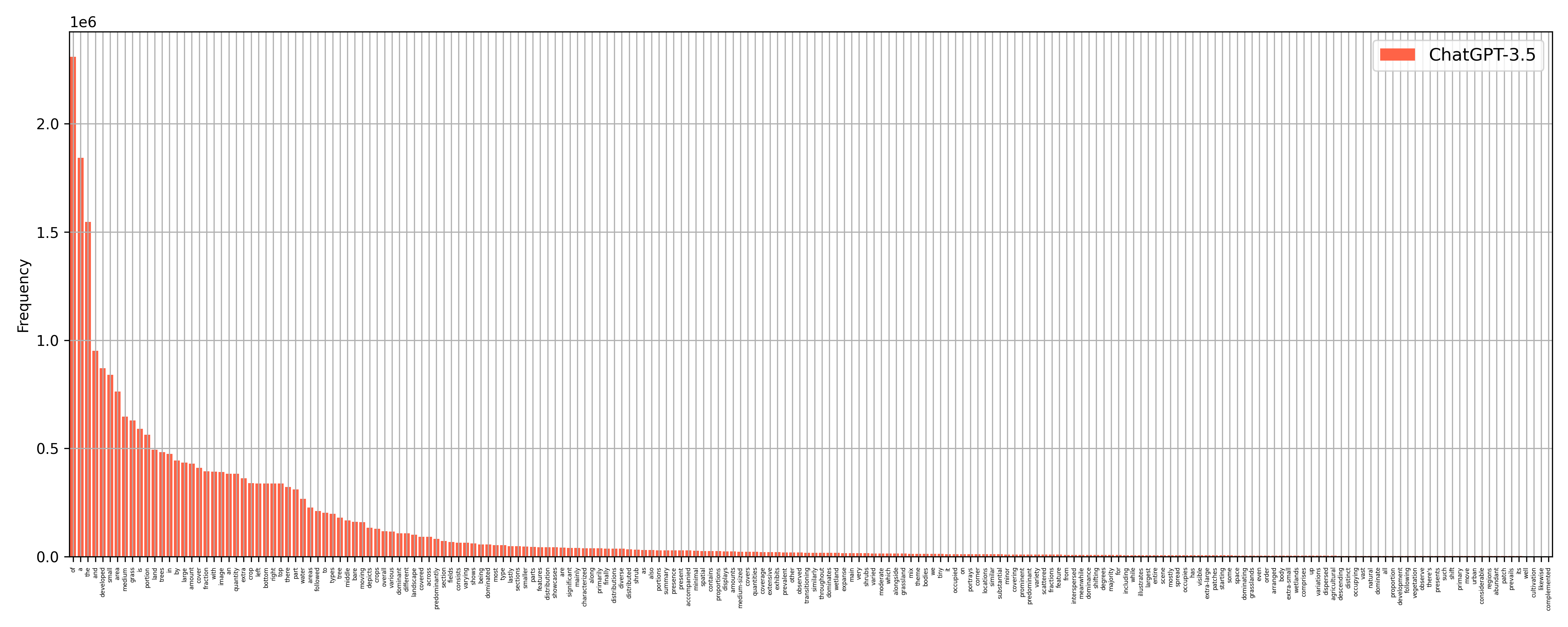}
\caption{Histogram of word frequencies for captions generated by ChatGPT-3.5.}\label{WF1}
\end{figure}

\begin{figure}[h]%
\centering
\includegraphics[width=0.99\textwidth]{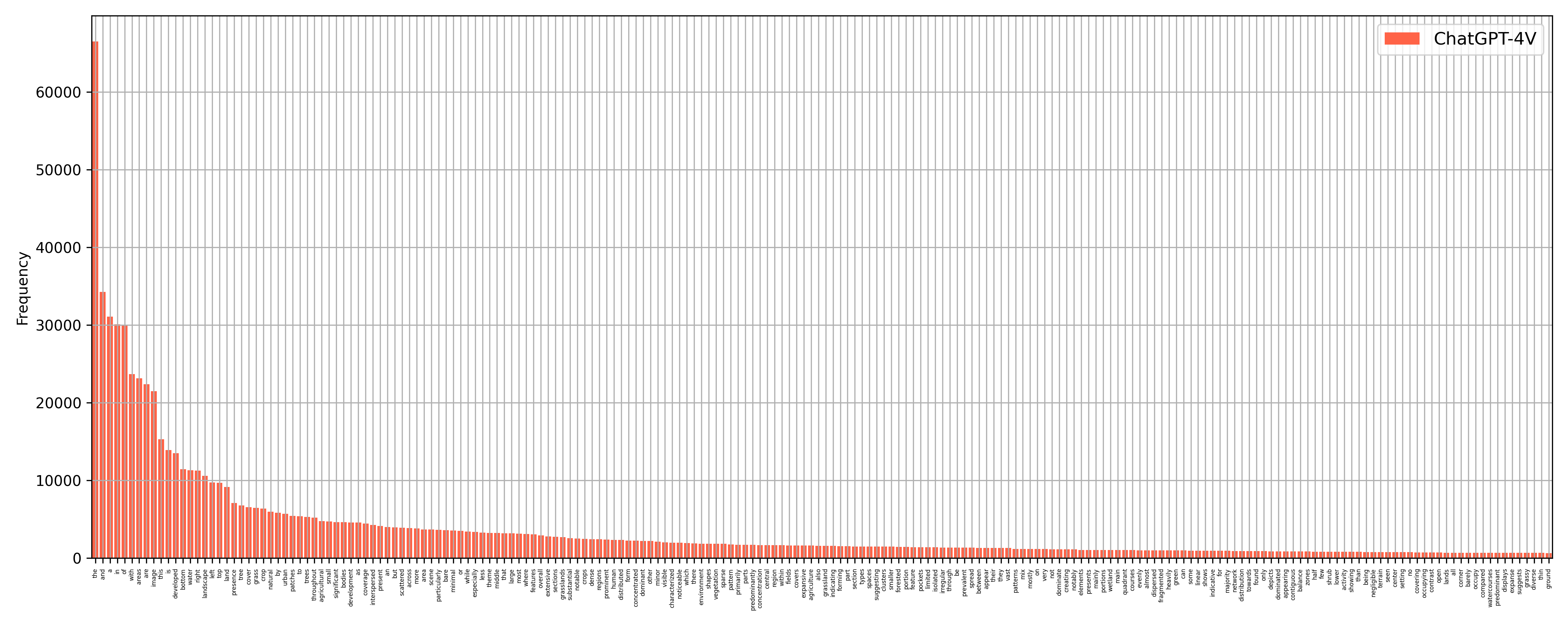}
\caption{Histogram of word frequencies for captions generated by ChatGPT-4V.}\label{WF2}
\end{figure}

Figure \ref{WF1} and Figure \ref{WF2} display histograms of the top 200 word frequencies for captions generated by ChatGPT-3.5 and ChatGPT-4V, respectively. The x-axis represents individual words, and the y-axis represents the frequency. Both distributions are long-tailed, indicating that a minority of words are used frequently, while the majority appear infrequently. High-frequency words such as ``the'', ``a'', ``of'', and ``and'' are predominantly present in both histograms due to their generic nature across textual data. Comparing two histograms, we observe that the descent from the most to less frequent words appears sharper in Figure \ref{WF1}, while Figure \ref{WF2} exhibits a more gradual decline. This observation indicates that ChatGPT-4V employs a broader vocabulary to generate more diverse and higher-quality captions.

\begin{figure}
  \begin{minipage}[t]{0.5\linewidth}
    \centering
    \includegraphics[scale=0.055]{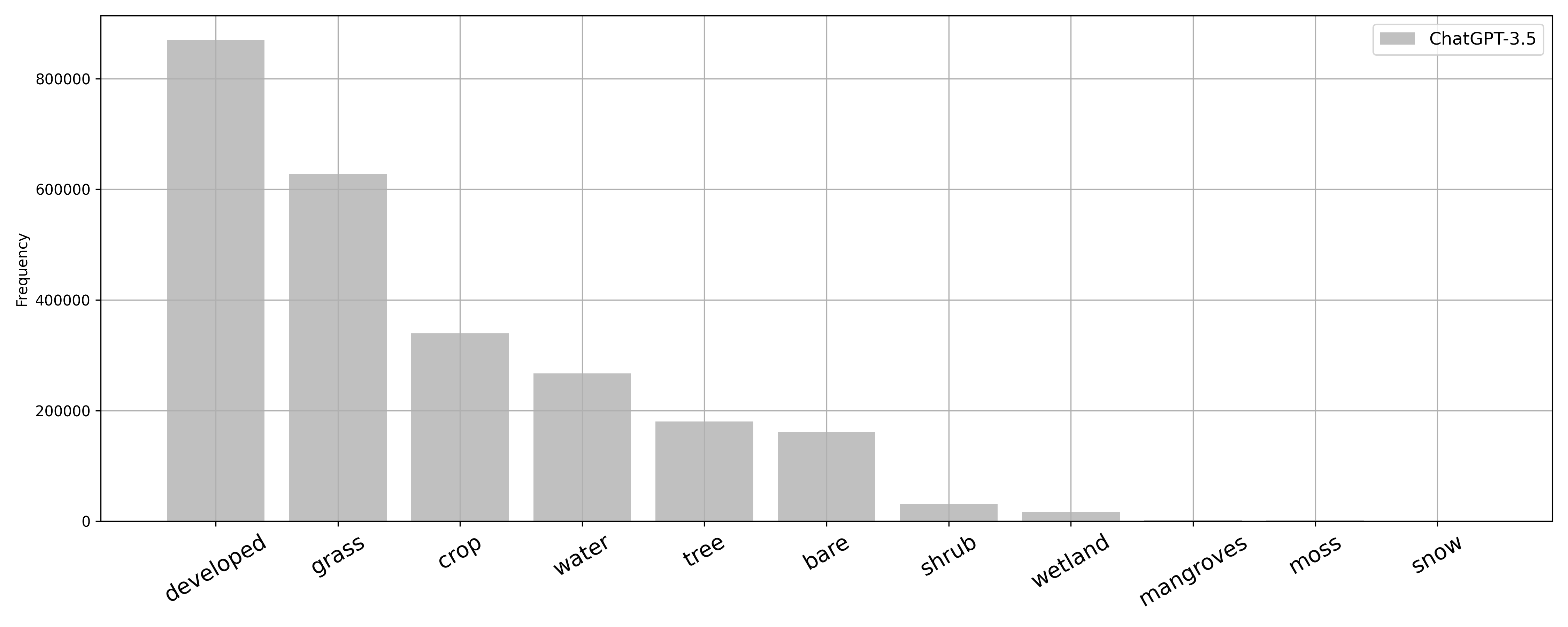}
    \caption{Histogram of word frequencies related to land cover \\types for captions generated by ChatGPT-3.5.}
    \label{stat_35_bar}
  \end{minipage}%
  \begin{minipage}[t]{0.5\linewidth}
    \centering
    \includegraphics[scale=0.055]{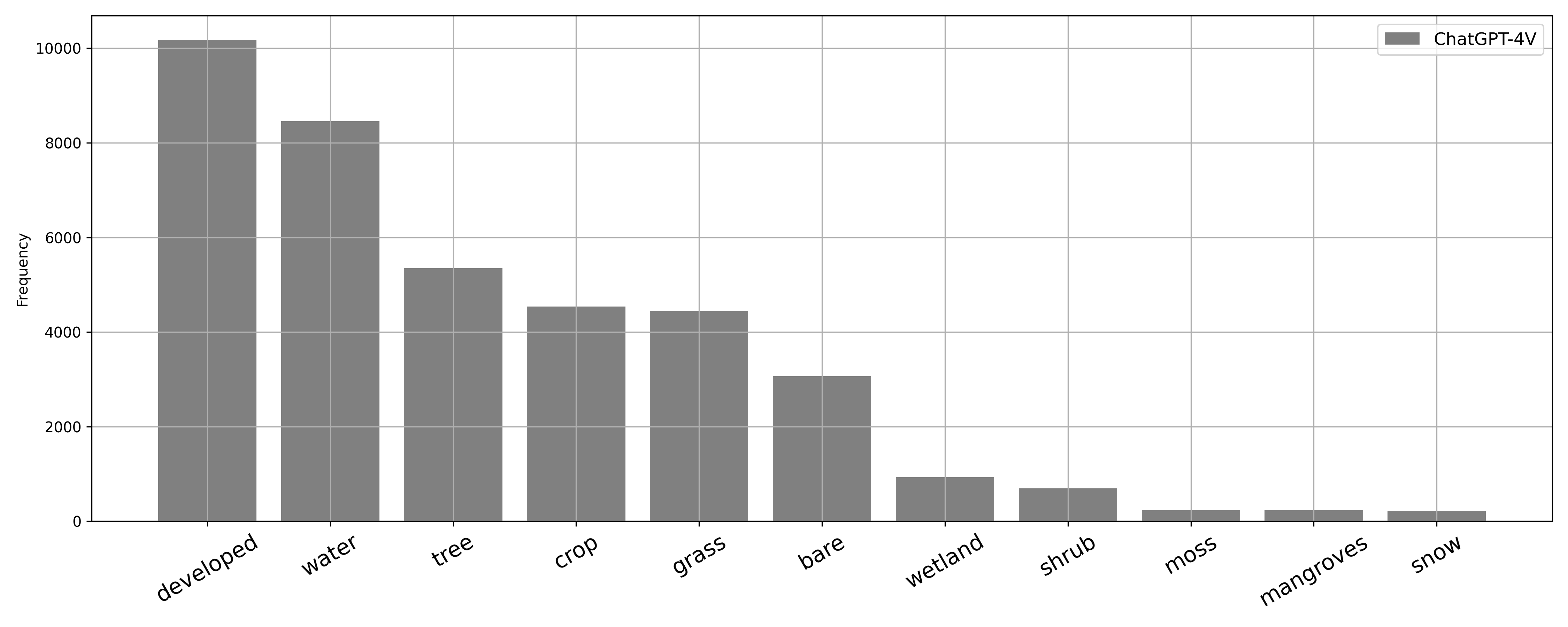}
    \caption{Histogram of word frequencies related to land cover \\types for captions generated by ChatGPT-4V.}
    \label{stat_4v_bar}
  \end{minipage}
\end{figure}

\begin{figure}
  \begin{minipage}[t]{0.5\linewidth}
    \centering
    \includegraphics[scale=0.055]{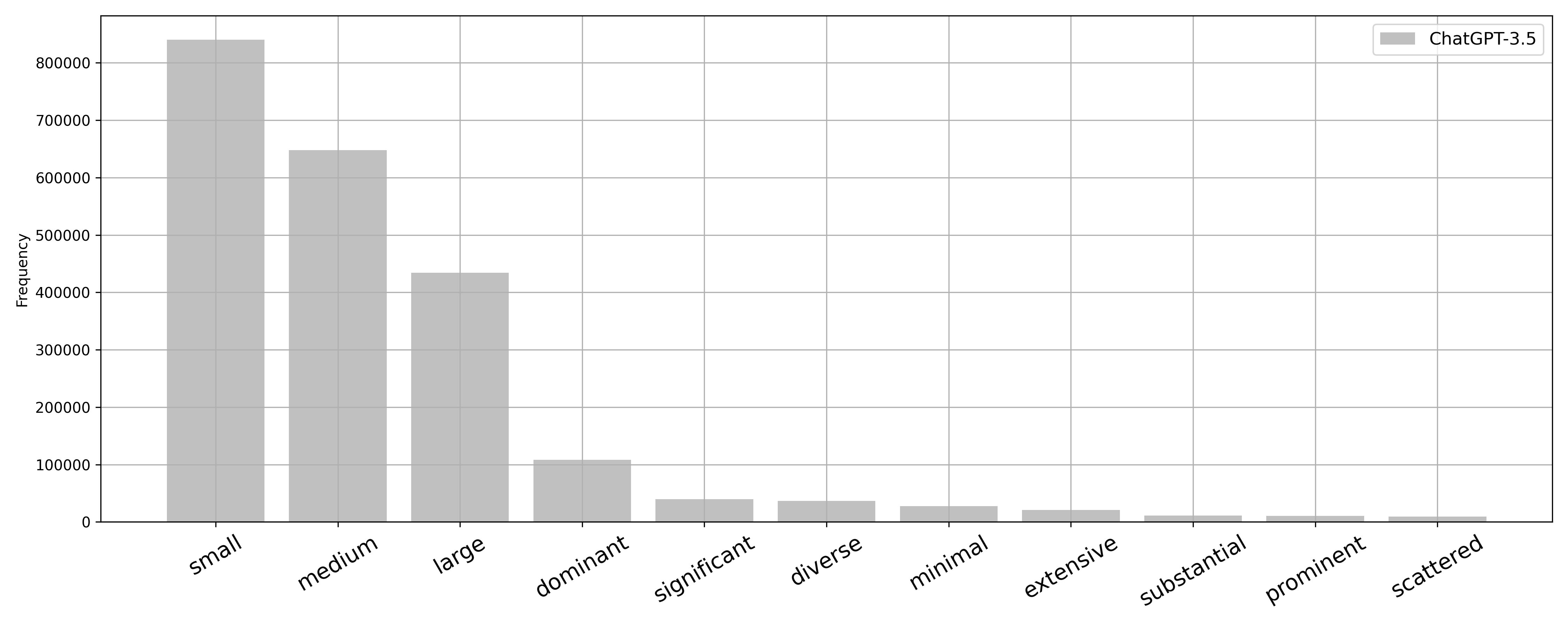}
    \caption{Histogram of word frequencies related to quantity and \\shape for captions generated by ChatGPT-3.5.}
    \label{stat_shape_35_bar}
  \end{minipage}%
  \begin{minipage}[t]{0.5\linewidth}
    \centering
    \includegraphics[scale=0.055]{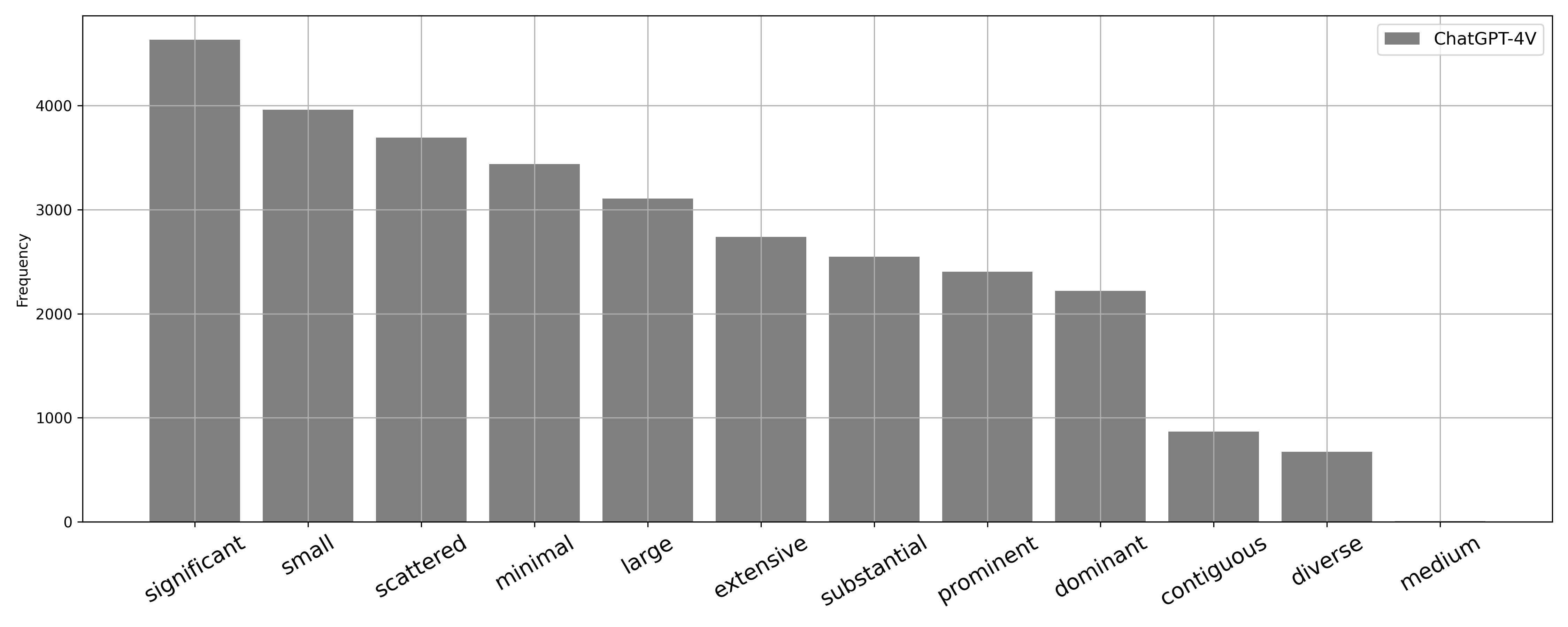}
    \caption{Histogram of word frequencies related to quantity and \\shape for captions generated by ChatGPT-4V.}
    \label{stat_shape_4v_bar}
  \end{minipage}
\end{figure}

To better understand the differences in captions related to land cover types generated by ChatGPT-3.5 and ChatGPT-4V, we construct histograms to illustrate the frequencies of relevant words, as depicted in Figure \ref{stat_35_bar} and Figure \ref{stat_4v_bar}. The x-axis represents land cover types, and the y-axis represents the frequency. The histogram in Figure \ref{stat_35_bar} exhibits a clear long-tailed distribution, with ``developed area'', ``grass'', and ``crop'' being the most frequently mentioned land cover types. In Figure \ref{stat_4v_bar}, ``developed area'', ``water'', and ``tree'' are predominant land cover types. These differences reflect the different descriptive approaches and varied geographical distributions in the two versions. Despite these differences, both versions display a largely similar distribution in terms of land cover types.

Figure \ref{stat_shape_35_bar} and Figure \ref{stat_shape_4v_bar} illustrate the word frequencies related to quantity and shape for two versions of captions generated by ChatGPT-3.5 and ChatGPT-4V, respectively. The x-axis represents words related to quantity and shape, and the y-axis represents the frequency. The descriptive terms between the two versions form a clear contrast. The histogram for ChatGPT-3.5 shows a preference for terms like ``small'', ``medium'', ``large'', and ``dominant'' to describe land cover proportions. Meanwhile, ChatGPT-4V, as reflected in the histogram, employs a more diverse vocabulary, extending beyond common descriptors such as ``small'', ``large'', and ``dominant'' to include high frequencies of ``significant'', ``scattered'', ``minimal'', ``extensive'', and ``substantial''. These words enrich the descriptions of land cover type shapes and patterns, indicating that captions of the ChatGPT-4V version leverage a broader vocabulary to describe the characteristics of the image.

\begin{figure}[h]%
\centering
\includegraphics[width=0.99\textwidth]{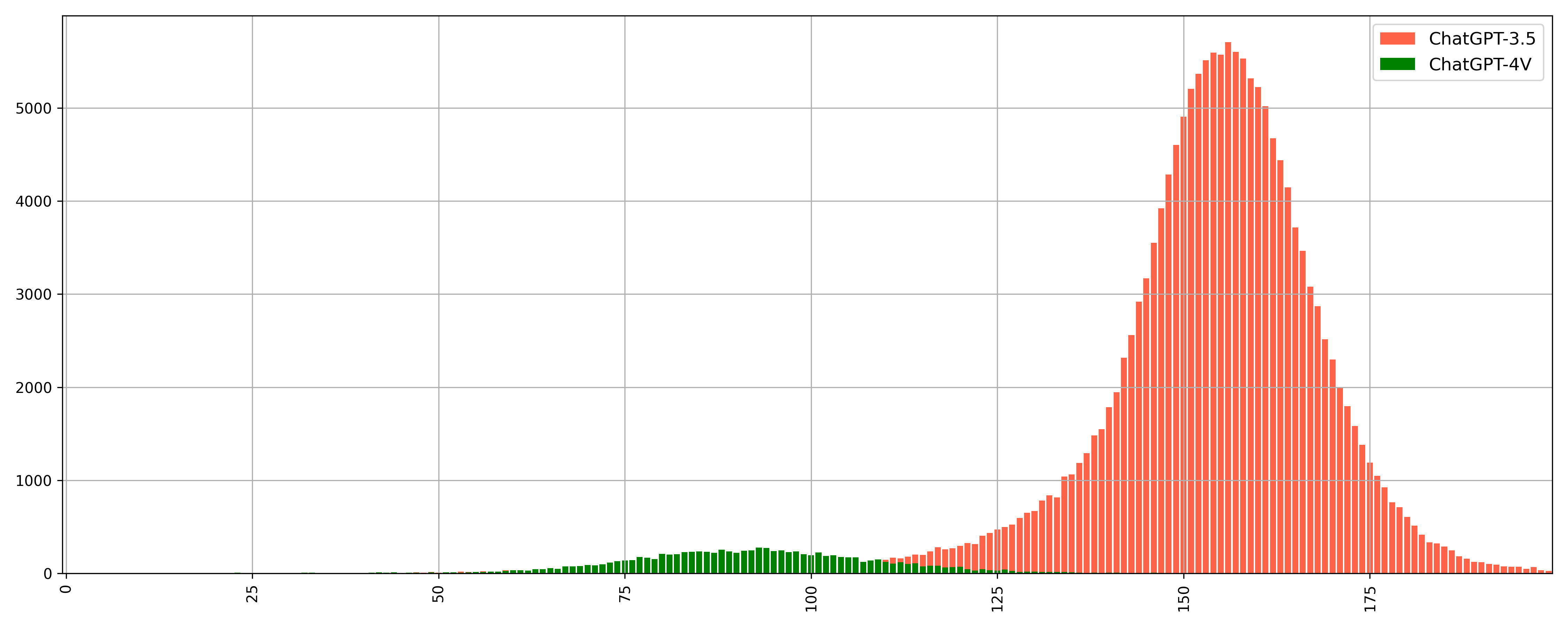}
\caption{Histogram comparing caption lengths generated by ChatGPT-3.5 and ChatGPT-4V.}\label{WLC}
\end{figure}

\subsection{Caption length}
Figure \ref{WLC} presents a comparison of caption lengths generated by ChatGPT-3.5 and ChatGPT-4V, illustrated as the histogram. The x-axis denotes caption length, and the y-axis represents the frequency of captions at each length. Unlike most existing image-text datasets that typically provide brief annotations, the ChatEarthNet dataset stands out by offering comprehensive captions that have detailed semantic insights into land cover types. The histogram for ChatGPT-4V, shown in green, forms a Gaussian distribution with a mean value of around 90 words per caption. The histogram for ChatGPT-3.5, depicted in orange, also shows a Gaussian distribution but with a mean centered around 155 words, suggesting that captions generated by this version are generally longer. The reason is that ChatGPT-3.5 tends to elaborate on provided prompts by extending contextual cues, resulting in detailed descriptions that try to encompass various aspects of prompts. Conversely, ChatGPT-4V comprehensively grasps contextual information in prompts, enabling it to generate concise yet comprehensive descriptions. Additionally, ChatGPT-4V harnesses visual data (land cover maps), to achieve a more precise comprehension of spatial distributions of land covers. As mentioned in Figure \ref{WF1} and Figure \ref{WF2}, captions in the ChatGPT-4 version utilize a more diverse vocabulary. Consequently, the ChatGPT-4V captions manage to be more concise yet more varied.

\begin{figure}[h]%
\centering
\includegraphics[width=0.99\textwidth]{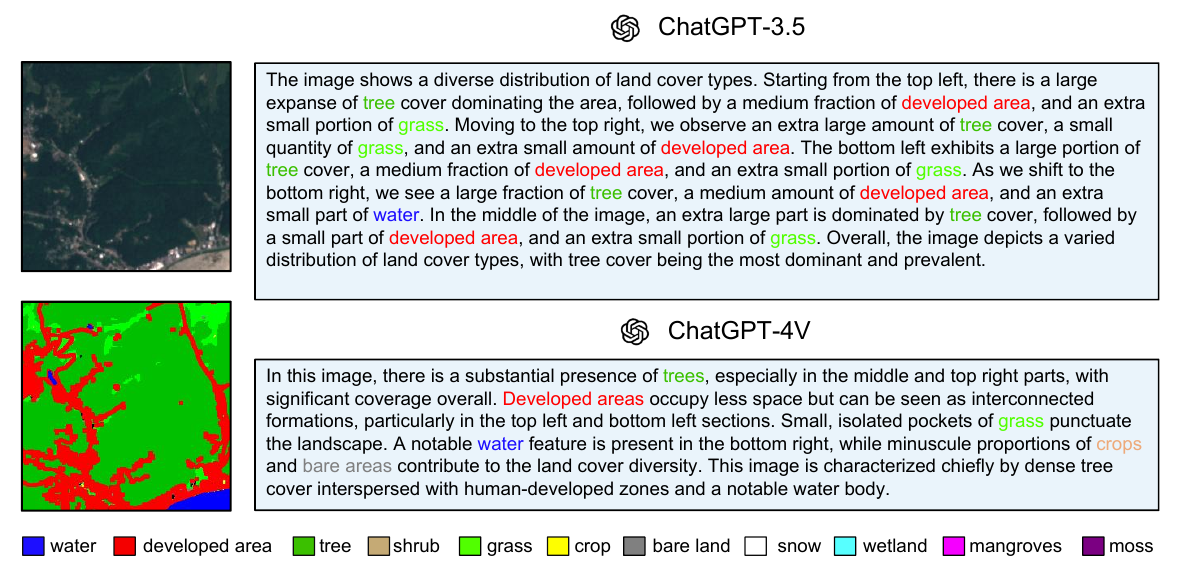}
\caption{Sentinel-2 satellite image, its associated land cover map, and its corresponding captions generated by ChatGPT-3.5 and ChatGPT-4V.}\label{Compare}
\end{figure}

\subsection{Visualization and Comparison}
In Figure \ref{Compare}, we showcase captions generated by ChatGPT-3.5 and ChatGPT-4V for a detailed comparison between the two versions. The caption from ChatGPT-3.5 provides a structured breakdown of the land cover types in five sections (top left, top right, bottom left, bottom right, and middle) of the image. This is a result of ChatGPT-3.5's inability to process image inputs directly, heavily relying on the given prompts. By doing so, these captions are structured, quantitative, and exhaustive, providing a balanced view of land cover types. In contrast,  the caption from ChatGPT-4V adopts a holistic perspective, depicting land covers in the context of the complete image rather than discrete sections. The language is descriptive and vivid,  emphasizing visually striking features and the general impression of the landscape. As ChatGPT-4V employs land cover maps as visual inputs, the generated captions offer a more comprehensive perspective, emphasizing the overall visual impact. While both captions offer different interpretations, each remains factually correct. The captions in the ChatEarthNet dataset can be valuable resources for the advancement of VLMs in the field of remote sensing.






\section{Conclusion}

In this work, we construct ChatEarthNet, a large-scale image-text dataset characterized by its global coverage, high quality, wide-ranging diversity, and detailed descriptions. Specifically, we utilize Sentinel-2 data for its global coverage as the image source, and we employ land cover maps from the ESA’s WorldCover project to generate text. Consequently, by analyzing these land cover maps, we are able to extract the spatial distributions of different land cover types, which serve as the context information for crafting the prompts. These well-curated prompts are employed to elicit descriptive captions for each Sentinel-2 image from two LLMs, ChatGPT-3.5 and ChatGPT-4V. ChatEarthNet comprises 163,488 image-text pairs with captions generated by ChatGPT-3.5 and an additional 10,000 pairs with captions generated by ChatGPT-4V. By combining the high-quality captions with the visual information from Sentinel-2 imagery, ChatEarthNet is a valuable resource for training vision-language geo-foundation models and evaluating large vision-language models for remote sensing.








\appendixfigures  

\appendixtables   










\bibliographystyle{copernicus}
\bibliography{ref.bib}

\end{document}